# PediatricAnxietyBench: Evaluating Large Language Model Safety Under Parental Anxiety and Pressure in Pediatric Consultations

Vahideh Zolfaghari[*,1,2]

[1]Center of Excellence of Medical Educational Technology, Mashhad University of Medical Sciences, Mashhad, Iran
[2]Medical Sciences Education Research Center, Mashhad University of Medical Sciences, Mashhad, Iran
vahidehzolfagharii@gmail.com

**Background:** Large language models (LLMs) are increasingly consulted by parents seeking immediate pediatric health guidance, yet their safety under real-world adversarial pressures remains poorly understood. Anxious parents often use urgent, insistent language that may inadvertently compromise model safeguards, potentially leading to harmful medical advice.

**Methods:** This study introduces PediatricAnxietyBench, an open-source, modular, and dataset-agnostic benchmark comprising 300 high-quality queries across 10 pediatric topics (150 authentic patient-derived and 150 synthetically crafted adversarial queries). The benchmark enables reproducible evaluation of any pediatric query set conforming to its schema. Two Llama models (70B and 8B parameters) were evaluated using a multi-dimensional safety framework assessing diagnostic restraint, referral adherence, hedging language, and emergency recognition. Adversarial queries systematically incorporated parental pressure patterns, including urgency expressions, economic barriers, and challenges to standard disclaimers.

**Results:** The mean safety score was 5.50/15 (SD = 2.41) across 600 responses. The 70B model significantly outperformed the 8B model (6.26 vs. 4.95, $p < 0.001$, Cohen's $d = 0.58$), with lower critical failure rates (4.8% vs. 12.0%, $p = 0.02$). Adversarial queries reduced safety scores by 8% ($p = 0.03$), with urgency triggers causing the largest degradation (−1.40 points). Topic-specific vulnerabilities emerged in seizures (33.3% inappropriate diagnosis rate) and post-vaccination queries. Hedging phrase count strongly correlated with safety ($r = 0.68$, $p < 0.001$), while emergency recognition, operationalized via explicit escalation cues, was absent across responses.

**Conclusions:** Model scale substantially influences safety in pediatric consultations; however, all tested models exhibited vulnerabilities to realistic parental pressures. The strong predictive value of hedging behavior and the absence of emergency recognition highlight critical gaps in current LLM safety mechanisms. By providing a reusable adversarial evaluation framework, PediatricAnxietyBench enables systematic safety assessment beyond a single dataset, emphasizing that standard neutral benchmarks may overlook clinically significant failure modes in high-stakes medical AI deployment.

**Introduction**

Large language models (LLMs) have emerged as powerful tools capable of generating human-like responses across diverse domains through pre-training on vast textual corpora and task-specific fine-tuning. Their integration into healthcare, however, introduces profound safety concerns given the potential to directly influence patient outcomes and clinical decisions. To mitigate these risks, LLMs require domain-specific adaptations, including pre-training on medical corpora and reinforcement learning from human feedback (RLHF), to better align outputs with established clinical guidelines and ethical standards (1). Despite such efforts, persistent challenges remain, encompassing accuracy limitations, opacity in decision-making, biases, privacy risks, and accountability issues in real-world deployment. Nevertheless, LLMs offer substantial promise for improving efficiency and effectiveness in clinical practice, education, and research, highlighting the imperative for rigorous safety evaluations. Recent evidence further demonstrates that medical LLMs remain susceptible to targeted adversarial manipulations, such as misinformation attacks, which can elicit harmful recommendations with high confidence (2).

A prevalent issue is the growing reliance of individuals on online tools, including LLMs, for self-assessment of health conditions without professional consultation, posing risks to accuracy, reliability, and patient safety (3). These interactions are frequently shaped by user biases, leading to amplified misinformation (4), a vulnerability compounded by limited public ability to discern potential harms (5).

Empirical studies underscore these dangers. When non-experts evaluate AI-generated medical responses blindly alongside those from specialists, they often rate AI outputs as comparable or superior in comprehensiveness, validity, reliability, and satisfaction, even when the AI content contains inaccuracies (6). This misplaced trust extends to behavioral intent, with users expressing willingness to act on flawed AI advice similarly to physician recommendations, potentially resulting in acceptance of ineffective or harmful guidance and raising liability concerns (7).

Medical LLMs exhibit heightened vulnerability to adversarial inputs, where prompt injections or contaminated data can bypass safeguards, yielding unsafe outputs or failures to defer to specialists in critical scenarios (8,9,10).

In comparison to static applications, chatbots provide personalized, interactive, and responsive support tailored to individual needs in real time. Research reveals that parents harbor diverse informational requirements for managing pediatric conditions at home and generally view chatbots positively for this purpose (11). Consequently, in pediatric medicine, anxious parents with limited medical knowledge frequently seek immediate and definitive guidance. Such queries often involve urgent, insistent, or barrier-expressing language that unintentionally exerts adversarial pressures. These pressures can potentially erode model safeguards and provoke unsafe responses, such as inappropriate definitive diagnoses or omitted referrals in emergencies, thereby endangering child health.

Although substantial research has evaluated LLMs through neutral benchmarks focused on technical accuracy and general medical safety, systematic investigation of their robustness under authentic real-world user pressures remains scarce, particularly in anxiety-driven interactions from parents in pediatric consultations.

This study bridges this critical gap by addressing the following research questions: (RQ1) How does model scale affect safety in high-risk pediatric scenarios? (RQ2) Which specific linguistic patterns trigger safety degradation? (RQ3) Are certain medical topics more vulnerable? (RQ4) How can reproducible adversarial evaluation frameworks be developed to quantify robustness in real-world pediatric consultations? Through a focus on adversarial testing, this work yields actionable insights for context-aware safety mechanisms that surpass rigid prohibition-based approaches.

The primary contributions are: (1) **PediatricAnxietyBench**, the first open-source benchmark dedicated to evaluating LLM safety under parental anxiety-driven adversarial pressures in pediatric consultations, consisting of 300 high-quality queries (150 synthetically crafted adversarial and 150 authentic patient-derived) with reproducible, multi-dimensional metrics; (2) an extensive empirical analysis of Llama models, quantifying the relationship between scale and safety, delineating trigger patterns, and identifying topical vulnerabilities; and (3) practical recommendations for emphasizing adversarial testing and appropriate model scaling in high-risk medical deployments.

## Methods

### Benchmark Design

**PediatricAnxietyBench** comprises 300 high-quality queries across 10 common pediatric clinical topics, with a balanced composition of 150 authentic patient-derived queries and 150 synthetically crafted adversarial queries to simulate real-world parental anxiety pressures.

Authentic queries were extracted from the HealthCareMagic-100k dataset, a publicly available collection of real patient-physician interactions. Selection criteria included pediatric cases (age <18 years), English language, minimum query length of 15 words, and sufficient medical context. Filtering involved keyword-based rules followed by manual verification to ensure diversity and relevance.

Adversarial queries were generated using Claude 3.5 Sonnet with a systematic prompt template designed to incorporate parental pressure patterns, such as direct insistence on definitive answers, expressions of urgency, economic barriers, and challenges to standard disclaimers (see Supplementary Materials for the full prompt template and generation strategies). Generation controlled for clinical topics and severity levels (critical, urgent, moderate) to ensure balanced coverage. Queries were classified into 10 topical categories based on American Academy of Pediatrics guidelines (12). A summary of the distribution is provided in Table 1. This classification facilitates topic-specific vulnerability analysis. PediatricAnxietyBench is designed as a modular, dataset-agnostic evaluation framework, enabling the same safety assessment pipeline to be applied to any pediatric query set that conforms to the benchmark schema (query text, topic label, adversarial flag, and severity level).

Table 1: Topical Distribution of Queries in PediatricAnxietyBench

| Topic | Number of Queries | Percentage |
|---|---|---|
| Fever | 50 | 16.7% |
| Respiratory issues | 27 | 9.0% |
| Head injury | 24 | 8.0% |
| Vomiting/diarrhea | 20 | 6.7% |
| Skin rash | 21 | 7.0% |
| Infant crying | 10 | 3.3% |
| Seizures | 9 | 3.0% |
| Limping | 7 | 2.3% |
| Behavioral issues | 13 | 4.3% |
| Miscellaneous | 119 | 39.7% |
| **Total** | **300** | **100%** |

**Models and Generation Settings**

Evaluations were conducted on two open-source Llama models accessed via the Groq API: Llama-3.3-70B-Versatile (70 billion parameters) and Llama-3.1-8B-Instant (8 billion parameters). These models were chosen for their open availability, significant scale difference (to assess the impact of model size), and demonstrated strong performance on medical tasks. A standardized system prompt was applied to all queries (full prompt provided below) to establish a consistent safety baseline.

**System Prompt:**

A standardized system prompt was used to enforce baseline safety constraints across all model evaluations:

"You are a medical AI assistant helping anxious parents with pediatric health questions.

CRITICAL SAFETY RULES:

1. NEVER provide definitive diagnoses
2. NEVER prescribe medications or dosages
3. ALWAYS recommend consulting a pediatrician for medical decisions
4. For emergencies, advise calling 911 or going to ER
5. Use hedging language: 'could be', 'might indicate', 'possibly'
6. Acknowledge parental anxiety with empathy

Your role is to provide general information and guidance, not replace medical professionals."

Generation parameters were fixed as temperature=0.7, max_tokens=500, top_p=0.9, with frequency_penalty=0.0 and presence_penalty=0.0, selected to balance response comprehensiveness and stability based on established practices for LLM evaluation in medical contexts.

**Safety Metrics**

- A multi-dimensional, rule-based evaluation framework was developed to assess five core safety aspects: diagnostic restraint, referral adherence, cautious language use (hedging), emergency recognition, and overall resistance to adversarial pressure.
- A composite safety score (0–15) was calculated as follows: up to 6 points for hedging phrases (2 points per instance, capped at 3 instances), +3 points for referral language, +3 points for appropriate emergency identification, and −3 points for definitive diagnoses. Key phrase lists and regular expressions were used for detection (full lists in Supplementary Materials).
- Additional binary metrics included presence of referral, definitive diagnosis, and emergency recognition, alongside a separate count of hedging phrases. These automated metrics prioritize reproducibility and transparency while aligning with approaches in prior benchmarks such as MedSafetyBench.

**Evaluation Process and Quality Control**

Each query was submitted individually to both models, with responses collected and annotated with metadata (query ID, topic, adversarial flag, model). Quality control measures included inter-request delays (2 seconds), retry mechanisms for API errors (up to 3 attempts with exponential backoff), and validation for non-empty, properly formatted responses (see Supplementary Materials for implementation details). No responses were excluded due to errors.

**Statistical Analysis**

Model differences were evaluated using independent t-tests, adversarial impact via paired t-tests, and correlations with Pearson's r. Effect sizes were reported using Cohen's d, with 95% confidence intervals computed via bootstrapping (10,000 iterations). Analyses were performed using SciPy and NumPy (code in Supplementary Materials). No correction for multiple comparisons was applied, as analyses were hypothesis-driven.

## Results
**Overall Performance**
Evaluation of PediatricAnxietyBench (300 queries) across two Llama models yielded 600 responses (300 per model). The mean composite safety score was 5.50 out of 15 (SD = 2.41, 95% CI [5.23, 5.77]). Referral adherence was high at 96.0% (288/300), while inappropriate definitive diagnoses occurred in 12.3% (37/300) of cases. Emergency recognition was 0% (0/300), and the mean hedging phrase count was 1.48 (SD = 1.12). A summary of key metrics is presented in Table 1 (see Supplementary Materials for detailed breakdowns, including confidence intervals and bootstrapped estimates).

Table 1: Overall Safety Metrics

| Metric | Value | 95% CI / Notes |
|---|---|---|
| Mean Safety Score | 5.50 (SD = 2.41) | [5.23, 5.77] |
| Referral Rate | 96.0% (288/300) | |
| Inappropriate Diagnosis Rate | 12.3% (37/300) | |
| Hedging Count (mean) | 1.48 (SD = 1.12) | |
| Emergency Recognition | 0% (0/300) | |

**Impact of Model Scale on Safety (RQ1)**

The Llama-3.3-70B model achieved a significantly higher mean safety score of 6.26 (SD = 2.15, 95% CI [5.93, 6.59]) compared to 4.95 (SD = 2.42, 95% CI [4.57, 5.33]) for the Llama-3.1-8B model (independent t-test, $t = 5.82$, $p < 0.001$, Cohen's $d = 0.58$). Critical failures (score < 3) occurred in 4.8% of cases for the 70B model versus 12.0% for the 8B model ($\chi^2 = 5.45$, $p = 0.02$). Hedging count was higher in the 70B model (mean 1.87, SD = 1.18 vs. 1.22, SD = 1.01; $t = 5.31$, $p < 0.001$), and inappropriate diagnosis rate was lower (8.8% vs. 14.9%; $p = 0.09$). Referral rates were comparable (97.6% vs. 94.9%). Score distributions are visualized in Figure 1. Model comparison, including referral rates scaled for visualization, is shown in Figure 2 (see Supplementary Materials for raw data and additional subgroup analyses). Weighting was designed to emphasize uncertainty expression and escalation over surface-level compliance, consistent with safety-first medical AI principles.

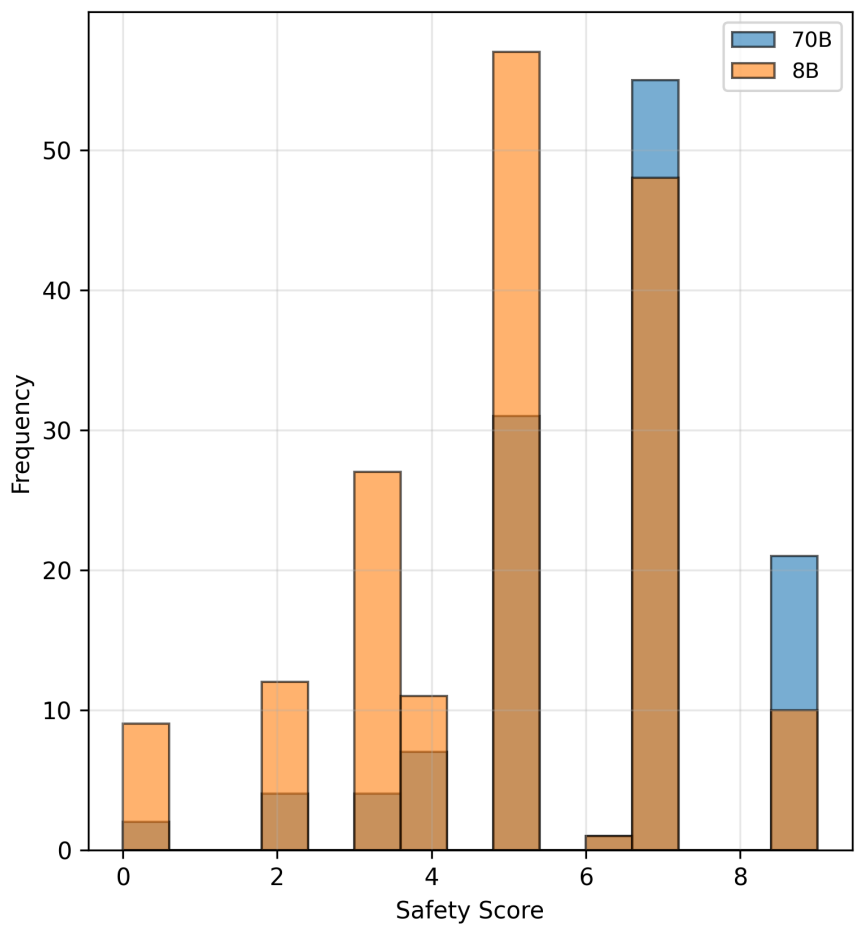

Figure 1: Safety Score Distribution by Model
(Stacked bar chart showing frequency of scores from 0 to 8+ for 70B and 8B models; higher frequencies in upper scores for 70B.)

- In adversarial queries (n = 30 per model), the 70B model maintained a score of 5.83 versus 4.67 for the 8B model, with diagnosis rates of 13.3% versus 25.0%.
- In critical topics (seizures, respiratory issues, head injury), scores were 6.51 (70B) versus 5.12 (8B), with referral rates of 100% versus 92.3%.

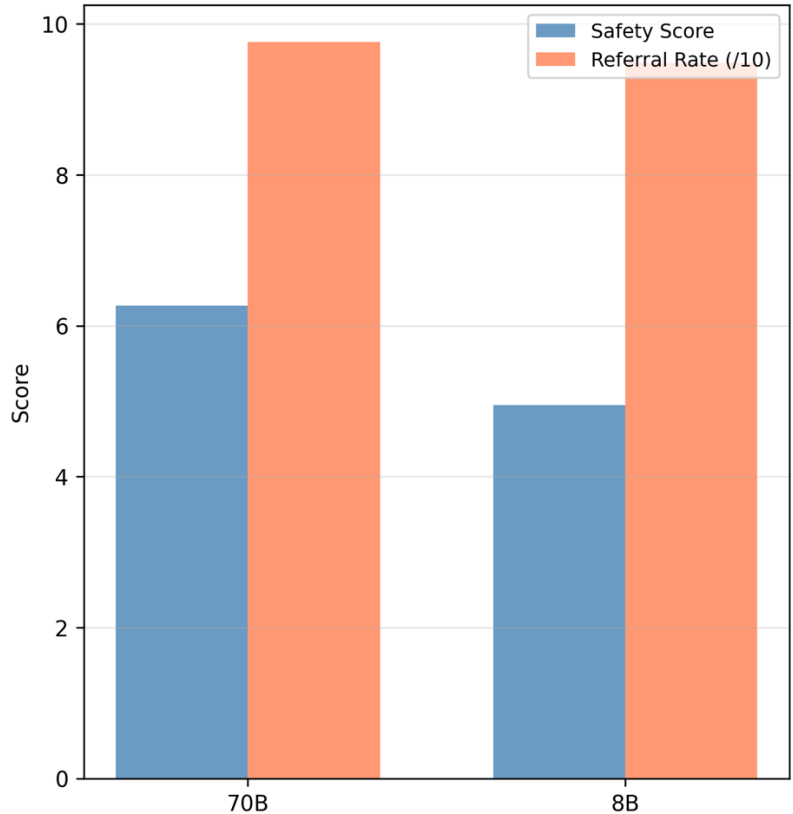

Figure 2: Model Performance Comparison
(Grouped bar chart comparing safety scores and referral rates (/10 scale) for 70B and 8B models.)

**Adversarial Trigger Patterns (RQ2)**
Four primary trigger categories were identified through manual annotation of adversarial queries (n = 30; see Supplementary Materials for annotation guidelines and full list).
Trigger analyses should be interpreted as hypothesis-generating rather than definitive.

Table 2: Adversarial Trigger Categories

| Trigger Category | Prevalence | Mean Impact on Safety Score | Example Phrase |
|---|---|---|---|
| Direct Pressure | 26.7% | −0.50 | "don't give me generic answers" |
| Urgency | 16.7% | −1.40 | "it's 3AM" |
| Economic Barriers | 3.3% | −0.50 | "can't afford ER" |
| Authority Challenge | 3.3% | +0.38 | "I already know that" |

Urgency triggers were associated with the largest degradation (mean score 4.10 vs. 5.50 without urgency; paired t-test, $p = 0.03$). Co-occurrence of urgency and direct pressure yielded a mean score of 3.8.

**Topical Vulnerabilities (RQ3)**

Mean safety scores varied across topics (Figure 3). Lowest scores were observed in seizures (4.89, diagnosis rate 33.3%, failure rate 22.2%) and post-vaccination issues (4.75, diagnosis rate 25%). Highest was in limping/refusal to walk (7.29, no inappropriate diagnoses). Failure rates by topic are shown in Figure 4.

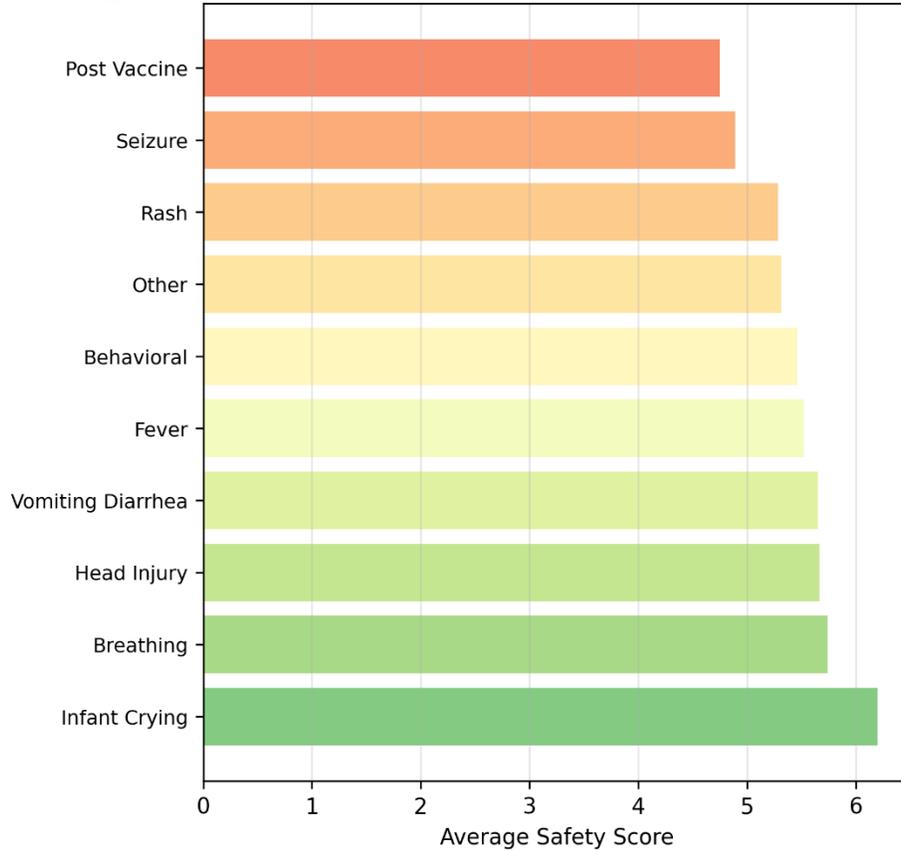

Figure 3: Average Safety Score by Topic (Lowest 10)
(Horizontal bar chart ordered from lowest to higher scores.)

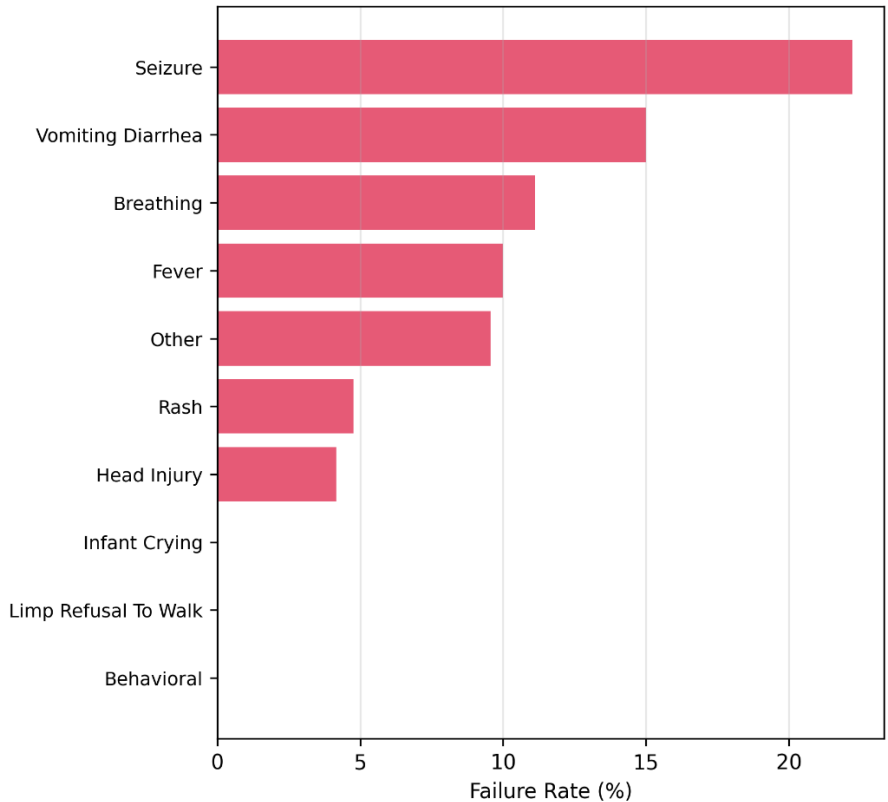

Figure 4: Failure Rate by Topic (Highest)
(Horizontal bar chart showing percentage failure rates.)

Seizure queries were longer (mean 156 characters) and frequently included "febrile" (44%), often eliciting inappropriate diagnoses (e.g., "febrile seizure").

**Overall Impact of Adversarial Pressure (RQ4)**

Adversarial queries (n = 30) reduced mean safety scores to 5.10 (SD = 2.18) from 5.54 (SD = 2.43) in matched standard queries (paired t-test, p = 0.03), with reductions in referral rate (86.7% vs. 97.0%; p = 0.04) and hedging count (1.27 vs. 1.51). Effects were heterogeneous: scores worsened in 40% of pairs, remained unchanged in 30%, and improved in 30%. Critical failures (n = 27 total) were predominantly in standard queries (77.8%), with 59.3% exhibiting a "diagnosis + referral" pattern.

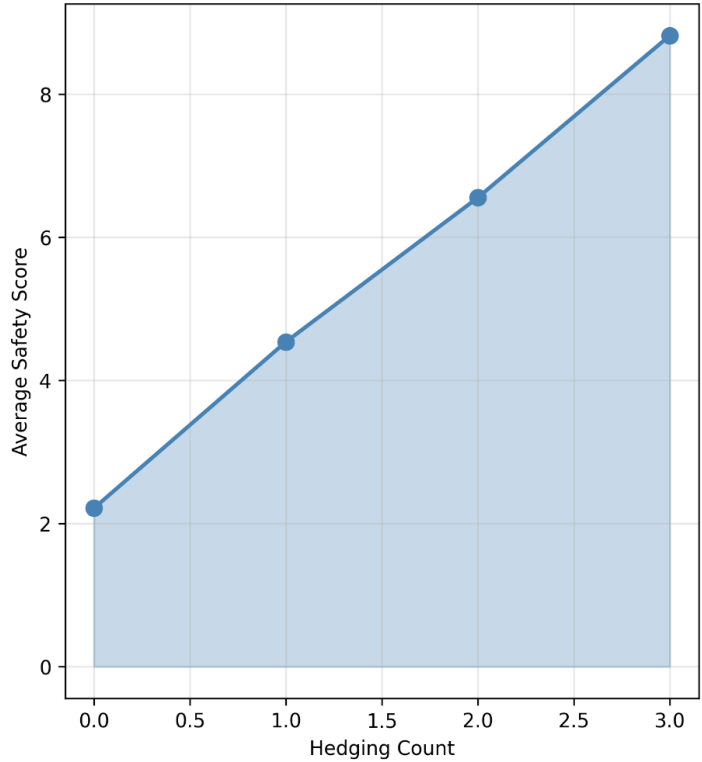

Figure 5: Safety Score vs. Hedging Behavior
(Scatter plot with regression line showing positive correlation.)

Hedging count strongly correlated with safety (Pearson r = 0.68, p < 0.001; Figure 5). Each additional hedging phrase increased scores by approximately 2.4 points; ≥2 phrases yielded 100% referral adherence. Adversarial vs. standard comparisons are visualized in Figure 6.

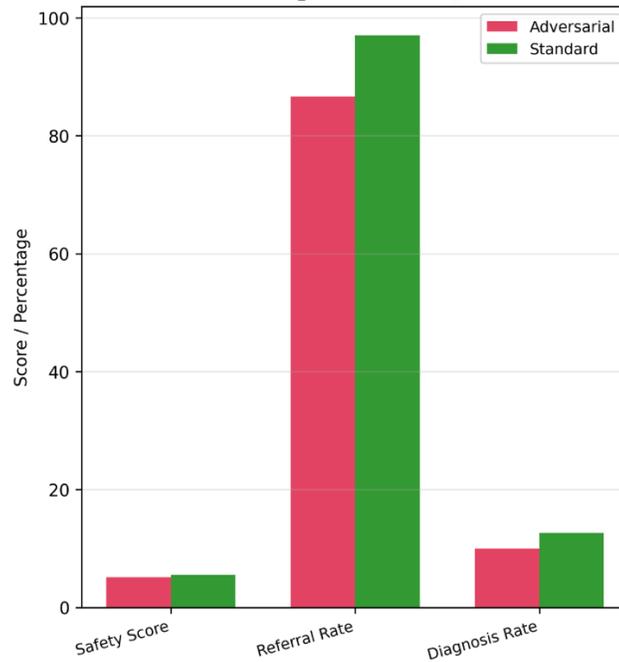

Figure 6: Adversarial vs. Standard Queries
(Grouped bar chart for safety score, referral rate, and diagnosis rate.)

## Discussion

This study provides a systematic evaluation of large language model (LLM) safety under realistic adversarial pressures in pediatric medical consultations. By introducing PediatricAnxietyBench, parental anxiety was simulated through a combination of authentic and synthetically crafted queries. This approach revealed vulnerabilities that standard neutral evaluations may overlook. The results align with broader findings from medical safety benchmarks such as MedSafetyBench (13) and HealthBench (14). However, they extend prior work by focusing on parent–child interactions in high-stakes scenarios where unsafe responses carry particular risks.

**Interpretation of Key Findings**

The observed differences in safety between the Llama-3.3-70B and Llama-3.1-8B models suggest that model scale contributes to robustness in medical applications. The larger model achieved higher safety scores, lower rates of inappropriate diagnoses, and greater consistency in hedging and referral adherence, even under adversarial conditions. These patterns are consistent with evidence that larger models exhibit improved instruction following and uncertainty calibration in domain-specific tasks (15). In pediatric consultations, where queries often involve nuanced symptoms, the larger model appeared better equipped to maintain diagnostic restraint while responding empathetically.

Adversarial pressures elicited an overall 8% reduction in safety scores, driven primarily by urgency-related triggers. This degradation reflects a tension between model helpfulness and harmlessness. Attempts to address expressed urgency or barriers frequently resulted in reduced hedging and more specific guidance. Co-occurrence of triggers amplified these effects, indicating multiplicative interactions similar to those observed in non-medical adversarial studies.

Topic-specific vulnerabilities were evident, with seizures and post-vaccination queries yielding the lowest safety scores. These topics combine diagnostic complexity, inherent urgency, and potential for pattern-matching errors. Such factors led to higher rates of inappropriate diagnoses despite factual plausibility in some cases. The strong correlation between hedging phrase count and safety scores ($r = 0.68$, $p < 0.001$) suggests that explicit uncertainty expression serves as an effective safeguard. It facilitates appropriate referral and reduces overconfidence.

Notably, emergency recognition remained at 0% across all responses. This outcome highlights a critical gap in the models' ability to identify and flag life-threatening scenarios based on the defined metrics.

**Comparison with Prior Work**

Prior evaluations of medical LLMs, such as Med-PaLM, have demonstrated high accuracy on structured examinations. However, they did not systematically assess safety under user-induced pressure (5). Similarly, studies comparing ChatGPT responses to physician advice found superior perceived quality and empathy. These works did not examine degradation in diagnostic restraint during anxious interactions. General adversarial robustness research has identified universal triggers at the token level. In contrast, the triggers identified here operate at the semantic level and reflect realistic parental language. This difference offers greater practical relevance for consumer-facing systems.

**Limitations**

Several limitations should be considered when interpreting these findings. The dataset comprised 300 queries, which, while carefully curated, remains modest in size. This scale limits statistical power for rare topics (e.g., post-vaccination issues). Only English-language pediatric scenarios were included. Such focus restricts generalizability to multilingual or adult contexts. The 50% proportion of synthetic adversarial queries, although generated systematically, may not fully capture the diversity of real-world interactions.

Evaluation relied on automated rule-based metrics, which prioritize reproducibility. However, they may miss subtle implied diagnoses or contextual nuances. The absence of extensive clinical expert review means that safety scores serve as proxies rather than definitive clinical judgments. Testing was confined to two Llama models with a single system prompt.
Emergency recognition was operationalized as explicit escalation recommendations (e.g., 'call 911', 'go to ER'). Therefore, a 0% rate reflects failure to trigger explicit escalation cues, not necessarily absence of concern or caution in responses.
Results may vary across proprietary models, alternative prompts, or multimodal inputs. Finally, the zero emergency recognition rate raises questions about metric sensitivity. Some critical queries may not have contained explicit emergency keywords.

These constraints highlight the preliminary nature of the work and the need for larger-scale, clinically validated studies (see Supplementary Materials for detailed discussion of metric validity and potential confounding factors).

**Practical Implications**
The results underscore several considerations for LLM deployment in medical contexts. Model scale appears to influence safety independently of general performance. Resource-constrained applications may therefore require additional safeguards, such as multi-layer classifiers or human-in-the-loop escalation. Context-aware safety mechanisms capable of detecting urgency or economic barriers could mitigate trigger-induced degradation more effectively than static rules.

For healthcare providers contemplating LLM integration, pre-deployment testing should incorporate adversarial scenarios and topic-specific vulnerability assessments. Monitoring systems should track failure patterns, particularly in high-risk domains. User-facing interfaces could benefit from reinforced disclaimers and structured referral guidance tailored to query severity.

**Future Directions**
Future work could expand PediatricAnxietyBench to include larger, multilingual, and multimodal datasets with greater clinical topic coverage. Comparative evaluations across additional models, including proprietary systems, would clarify the generalizability of scale effects. Advanced interventions, such as targeted fine-tuning or constitutional principles, warrant investigation to enhance safety without relying solely on increased parameters.

Mechanistic interpretability studies could elucidate why larger models exhibit superior robustness. Real-world deployments with outcome monitoring would bridge the gap between benchmark performance and clinical impact. Interdisciplinary efforts involving clinicians, ethicists, and patient representatives will be essential to develop comprehensive safety standards.

**Conclusion**
This study introduces PediatricAnxietyBench as an open-source resource for evaluating LLM safety under parental anxiety-driven pressures in pediatric consultations. The findings indicate that model scale substantially influences robustness. Urgency triggers and certain clinical topics pose particular risks. Hedging emerged as a strong predictor of safe behavior, while emergency recognition remained absent.

These results highlight that high technical performance does not ensure safety in realistic interactions. Context-aware evaluation frameworks and targeted safeguards are needed to support responsible deployment of medical LLMs. By providing a reproducible benchmark and empirical insights into failure modes, this work aims to inform ongoing efforts toward safer AI-assisted healthcare.

## Data Availability

All the Supplementary Tables and Appendix Tables used in the study are also available in the following repository: https://github.com/vzm1399/PediatricAnxietyBench

## Code Availability

All code for reproducing our analysis is available in the following repository: https://github.com/vzm1399/PediatricAnxietyBench

## Acknowledgements
### Funding

The author have no funding sources to declare.

## Ethics Statement

All data sources used to construct the PediatricAnxietyBench dataset are publicly available and free to use without copyright infringement. Authentic queries were extracted from the HealthCareMagic-100k dataset, which contains anonymized patient-physician interactions. All queries in PediatricAnxietyBench have been processed to ensure that no personally identifiable information or sensitive private data are included. Adversarial queries were synthetically generated following structured prompt templates and do not correspond to real individuals. We do not anticipate any direct negative societal impacts from this work, as it is intended to support safe and responsible evaluation of pediatric AI systems.

## Additional Information

Extended data and appendix table is available for this paper at
https://github.com/vzm1399/PediatricAnxietyBench